# Adaptive Spiking with Plasticity for Energy Aware Neuromorphic Systems

Eduardo Calle-Ortiz, Hui Guan, Deepak Ganesan, Phuc Nguyen
University of Massachusetts Amherst
ecalleortiz@umass.edu,huiguan@cs.umass.edu,dganesan@cs.umass.edu,vp.nguyen@cs.umass.edu

## ABSTRACT

This paper presents **ASPEN**, a novel energy-aware technique for neuromorphic systems that could unleash the future of intelligent, always-on, ultra-low-power, and low-burden wearables. Our main research objectives are to explore the feasibility of neuromorphic computing for wearables, identify open research directions, and demonstrate the feasibility of developing an adaptive spiking technique for energy-aware computation, which can be game-changing for resource-constrained devices in always-on applications. As neuromorphic computing systems operate based on spike events, their energy consumption is closely related to spiking activity, i.e., each spike incurs computational and power costs, consequently, minimizing the number of spikes is a critical strategy for operating under constrained energy budgets. To support this goal, **ASPEN** utilizes stochastic perturbations to the neuronal threshold during training to not only enhances the network's robustness across varying thresholds, which can be controlled at inference time, but also act as a regularizer that improves generalization, reduces spiking activity, and enables energy control without the need for complex retraining or pruning. More specifically, **ASPEN** adaptively adjusts intrinsic neuronal parameters as a lightweight and scalable technique for dynamic energy control without reconfiguring the entire model. Our evaluation on neuromorphic emulator and hardware shows that **ASPEN** technique significantly reduces spike counts and energy consumption while maintaining accuracy comparable to state-of-the-art methods.

## 1 INTRODUCTION

Wearable devices, which have become increasingly ubiquitous with over 500 million units shipped in 2024 alone [1], are now widely used for continuous, real-time monitoring in applications such as health tracking, rehabilitation, and human activity recognition. These systems integrate a variety of sensors capable of capturing more than 20 high-resolution physiological signals, including inertial measurement units (IMUs), optical sensors, and bioelectrical interfaces, producing multi-channel data streams with fine temporal granularity. Performing real-time inference on such data requires frequent activation of compute blocks, memory access, and signal transformation, all of which impose significant energy demands. IMU-based activity recognition, for example, involves constant sampling of tri-axial acceleration and angular velocity signals, followed by feature extraction and inference at sub-second intervals to capture user motion dynamics with low latency. Yet, these devices typically operate on battery capacities of less than 100 mAh, and in many use cases, must remain active for weeks to months on a single charge. This imposes a hard constraint on average system power consumption, often requiring sustained operation below one milliwatt. As a result, wearable AI systems face a core trade-off between inference accuracy, latency, and long-term energy sustainability. Achieving intelligent always-on sensing under such extreme energy constraints remains an open systems challenge.

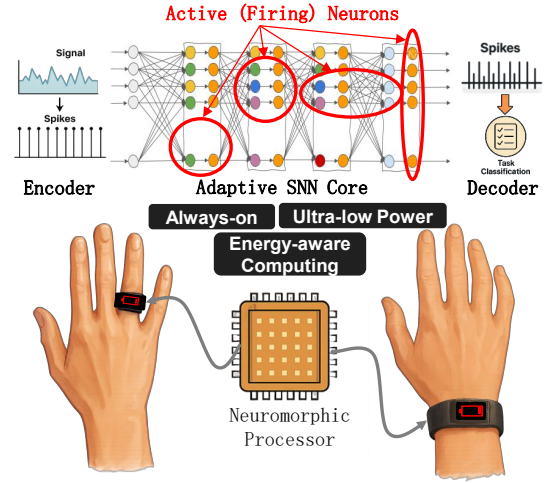

**Figure 1: ASPEN's concept**

A common strategy to address energy limitations is to offload sensing data to the cloud. While this reduces the computational burden on the device, it introduces high and variable latency, consumes significant wireless transmission energy, and raises privacy concerns [19, 23]. Traditional edge computing attempts to localize inference using optimized neural networks, but still relies on continuous, clock-driven computation. Techniques such as network pruning, quantization, neural architecture search, and knowledge distillation reduce computational complexity, but none are fundamentally designed for the ultra-low-power regimes required in always-on sensing scenarios. Moreover, these approaches are static and do not adapt to fluctuations in available energy, which are common in battery-constrained or energy-harvesting environments.

To overcome these limitations, neuromorphic systems have emerged as a promising architectural alternative to address the fundamental mismatch between continuous sensing demands and strict energy constraints in wearable devices. Unlike conventional deep learning architectures that rely on dense, clock-driven computation, neuromorphic systems adopt an event-driven model inspired by biological neural processing. These systems communicate using discrete electrical pulses, or spikes, and activate computation only in response to informative input events. This paradigm supports deep idle states and minimizes baseline power consumption, often achieving sub-microwatt operation levels [12]. Spiking Neural Networks (SNNs), the computational backbone of neuromorphic processors, leverage sparse temporal dynamics and operate using local memory and compute, making them especially well-suited for low-power, always-on tasks in constrained environments. Recent neuromorphic hardware platforms, such as Intel's Loihi 2 and SynSense's Xylo-IMU, demonstrate the potential of this approach by delivering up to 100× reductions in energy per inference compared to conventional CPU or GPU systems [5, 41]. These platforms are particularly attractive for wearable sensing applications, such as motion classification using IMUs, where spike-based signal encoding aligns well with the intermittent and structured nature of physical movement.

■ **Limitations of Current Neuromorphic Systems.** However, most existing SNN implementations employ static neuron configurations, with fixed parameters such as firing thresholds, membrane time constants, or refractory periods [34, 35]. These parameters are typically selected during training or manually tuned for a specific deployment scenario, and cannot adapt dynamically to changing energy availability or application demands. While these systems exploit data sparsity, their inability to modulate computational effort at runtime limits their responsiveness and energy scalability, particularly in wearable devices targeting months-long operation under variable conditions. This static nature prevents neuromorphic systems from realizing their full potential for adaptive, energy-aware computing in real-world deployment scenarios.

■ **ASPEN's Approach.** In this paper, we introduce **ASPEN**, a novel energy-adaptive neuromorphic computing technique designed to maintain always-on inference in wearable systems operating at microwatt-level power. As illustrated in Fig.1, **ASPEN** supports continuous sensing and inference on bio-inspired activity recognition tasks and is suitable for compact wearable platforms such as rings, ear-worn, or wrist-mounted devices.

Drawing from biological observations of intrinsic plasticity, where neurons adjust their excitability to maintain homeostasis and respond to varying stimuli [29], **ASPEN** introduces stochastic perturbations to neuron firing thresholds during training. This approach encourages networks to learn robust internal representations across a range of spiking conditions. At deployment time, **ASPEN** allows real-time control over energy usage by dynamically adjusting spiking activity without requiring model retraining or architectural modification.

The key innovation of **ASPEN** lies in its ability to integrate seamlessly with standard SNN training pipelines while introducing negligible computational overhead and supporting efficient deployment across neuromorphic hardware platforms. This enables runtime adaptability in energy-constrained wearables and supports continuous operation across diverse usage scenarios and energy profiles, effectively bridging the gap between the theoretical promise of neuromorphic computing and its practical deployment in resource-constrained environments.

■ **Challenges.** Realizing an adaptive, energy-aware control mechanism in spiking neural networks faces multiple technical challenges:

- *Learning robust threshold dynamics for post-training energy adaptation.* Spiking neural networks often rely on fixed threshold settings, which limits their adaptability and performance under energy constraints. However, adjusting neuron thresholds at inference time can destabilize network dynamics or degrade performance if not properly regularized during training. Learning threshold dynamics that are robust to variations in input conditions and energy budgets remains an open problem.

- *Predictable control of energy–accuracy trade-offs.* The relationship between threshold values, spike rates, and model accuracy is nonlinear and difficult to characterize/model. Small changes in threshold can produce unpredictable network behavior, especially when thresholds were not part of the training dynamics. A key challenge is to enable reliable and interpretable threshold modulation that provides calibrated control over energy and performance trade-offs.

- *Lack of adaptability across operating conditions.* Most existing neuromorphic models are trained and evaluated under fixed configurations, using static intrinsic parameters such as firing thresholds or membrane time constants. This makes it difficult to adjust model behavior in response to changing runtime conditions or energy availability. Current methods offer limited visibility into how accuracy, latency, and spike rate vary across different configurations. There is no support for post-training exploration or runtime adjustment *without retraining or modifying* the network architecture.

- *Hardware constraints on implementing adaptive behavior.* Deploying adaptive SNNs on neuromorphic platforms remains a major challenge due to hardware immaturity and limited configurability. Most platforms do not support runtime adjustment of neuronal parameters such as thresholds, and impose strict constraints on memory capacity, compute throughput, and numerical precision. Enabling energy-aware adaptation within these limitations requires techniques that operate within fixed hardware resources, avoid added overhead, and remain compliant with stringent power and latency budgets.

■ **Contributions.** This paper addresses the above challenges and makes the following technical contributions. First, we introde energy-aware spiking neural network learning models that, for the first time, enable the system to optimize energy consumption by directly regulating spiking activity within the network, thereby meeting strict energy budgets while maintaining sufficient accuracy for downstream tasks. Second, we introduce a novel bio-inspired approach that leverages intrinsic plasticity mechanisms to create self-adaptive neuromorphic systems capable of runtime energy optimization without requiring model retraining or architectural changes. Third, we develop a theoretical framework explaining how intrinsic plasticity can substantially reduce spike counts during inference, demonstrating the generalizability and scalability of the approach. Finally, we design, implement **ASPEN**, and evaluate on a widely used IMU-based dataset and neuromorphic hardware. The results showed that **ASPEN** reduces energy consumption to approximately 120 $\mu$W, which means around 1000× lower power than traditional von Neumann-based embedded systems and 2× lower than state-of-the-art neuromorphic systems such as SynSense-IMU. A 100 mAh, 3.7 V battery, such as those used in the Xiaomi Mi Band 3, Samsung Galaxy Fit, and Mi Smart Band 5, can power **ASPEN**'s IMU for approximately 128 days (or about 4.2 months) of continuous operation. In comparison, a 22 mAh battery, as found in the Oura Ring, can support always-on activity recognition for up to 22 days. It is important to note that none of these commercial devices support continuous, always-on gesture recognition; instead, they operate intermittently to conserve power.

In the following discussion, we provide background on neuromorphic computing and energy optimization via extrinsic and intrinsic plasticity (Section 2), introduce our intrinsic plasticity–based approach for building *ASPEN* (Section 3), detail its design (Section 4) and implementation on Rockpool and Xylo (Section 5), and present evaluation results and conclusions (Sections 6–7).

## 2 PRELIMINARIES

We now provide background on a new bio-inspired computational method for embedded systems, particularly wearables, called neuromorphic computing, with a focus on spiking neural network architectures for enabling ultra-low-power machine learning.

### 2.1 Bio-inspired Computation: Neuromorphic Computing

Traditional computing systems are based on the Von Neumann architecture, which separates memory from processing and depends on synchronized, clock-driven operations. While this model has supported decades of digital progress, including modern artificial intelligent systems, it remains inefficient in key areas, particularly in energy consumption and its limited ability to handle parallel information processing. In contrast, biological brains operate asynchronously, process information locally, and consume only a fraction of the energy, even when performing complex tasks such as vision, motion, or decision-making.

■ **What makes the biological brains so efficient?** Computational neuroscience has identified several principles that account for the brain's efficiency in computation, low latency, and minimal energy use [2, 38]. First, neurons and synapses function as both processing and memory units [16], which avoids the data transfer bottlenecks associated with conventional architectures that separate memory and computation. Second, the brain operates in a distributed and asynchronous fashion [10, 16]; neurons activate only when stimulated, reducing unnecessary computations and allowing faster response times with lower energy consumption. Third, biological systems encode spatial and temporal features of sensory inputs using sparse, spike-based signaling [26], which reduces communication overhead and energy demands. Finally, the brain's hierarchical organization [9, 27] enables it to solve complex tasks efficiently with relatively few processing stages, delay and power requirements.

■ **Capability to operate at ultra-low power.** Unlike conventional architectures, neuromorphic systems integrate computation and memory locally and operate in an event-driven, asynchronous manner. By mimicking how biological neurons communicate through spikes and adapt over time, neuromorphic systems offer a promising alternative to more sustainable computing. One of the most transformative advantages of neuromorphic systems lies in their ability to operate at extremely low power levels, often in the range of tens to hundreds of microwatts. This results in a significant reduction in energy consumption compared to conventional embedded processors, which typically consume power in the milliwatt to watt range even during low-power operation. The efficiency of neuromorphic systems is primarily due to their asynchronous, event-driven nature, where computation occurs only in response to input spikes and idle power consumption is negligible. For instance, neuromorphic chips such as Intel's Loihi and SynSense's DYNAP-SE have demonstrated energy costs of less than 23 $pJ$ per spike, allowing inference tasks to be executed within sub-$mW$ budgets.

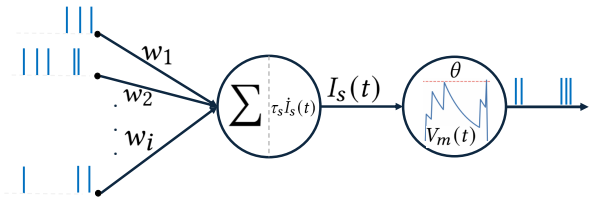

Figure 2: Leaky Integrate and Fire Model

At these power levels, continuous operation becomes feasible for long durations using compact energy sources. A neuromorphic system consuming 100 microwatts can operate for more than six months on a standard 200 mAh coin cell battery, assuming typical sensing and processing duty cycles. When coupled with energy harvesting technologies such as photovoltaic cells, thermoelectric generators, or piezoelectric harvesters, these systems can achieve long-term or even perpetual operation without manual battery replacement. This level of energy efficiency is especially valuable for wearable applications, where physical access is limited and battery replacement is burdensome. By aligning computational demands with stringent power and form-factor constraints, neuromorphic computing enables a new class of intelligent, maintenance-free wearable systems for continuous real-world deployment.

### 2.2 Neuromorphic Learning

■ **Spiking Neural Networks (SNNs).** At the core of many neuromorphic systems are SNNs, which emulate the discrete, event-driven communication observed in biological neurons. Unlike conventional artificial neural networks (ANNs) that rely on continuous-valued activations and synchronous processing, SNNs transmit information through sparse binary spikes, activating only when a neuron's membrane potential exceeds a threshold. This enables high temporal resolution, low average power consumption, and efficient asynchronous computation, making SNNs particularly suitable for real-time, sequential tasks such as auditory processing, gesture recognition, and physiological signal analysis (e.g., ECG, EEG, EMG) in wearable systems. Recent advances in neuromorphic hardware and algorithm co-design have demonstrated sub-milliwatt operation through specialized components like memristors [11, 24], while algorithmic innovations continue to improve SNN training and energy efficiency [36, 39]. Similar to the brain, the efficiency of SNNs arises from a parallel architecture combined with sparse, spike-driven processing, which reduces both computation and power requirements. These properties make neuromorphic computing uniquely suited for ultra-low-power, always-on, and even battery-free devices that must operate reliably under variable energy conditions.

■ **Leaky Integrate-and-Fire model (LIF) Model.** Different models for spiking neurons have been used to describe the dynamics of biological neurons. The most common spiking neuron model is the LIF model (Fig. 2) [14], where the dynamics of the neuron are

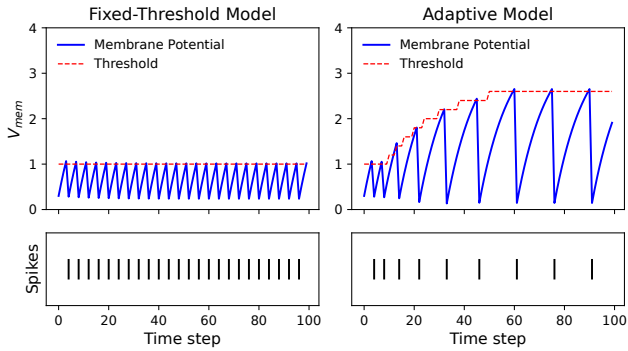

**Figure 3: Fix-Threshold vs Adaptive Threshold.**

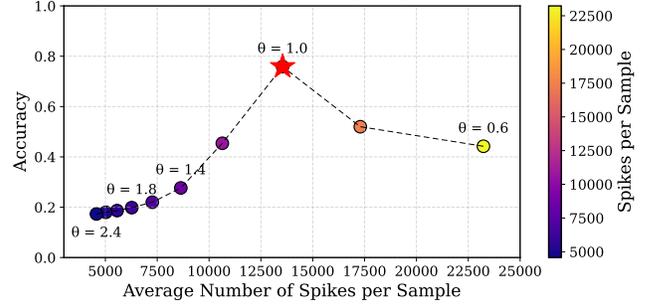

**Figure 4: Sensitivity Analysis Over Threshold Space in a Fixed-Threshold Model**

captured by the following equations:

$$\tau_s \frac{dI_s(t)}{dt} + I_s(t) = \sum_i w_i \sum_j \delta(t - t_j^i) \quad (1)$$

$$\tau_m \frac{dV_m(t)}{dt} = -V_m(t) + I_s(t) + b \quad (2)$$

$$s(t) = H(V_m(t) - \theta) \quad (3)$$

The input signal $I_s(t)$ (Fig.2) consists of weighted spikes from connected neurons, integrated over time using an exponential decay function with a time constant $\tau_s$. The time constant sets how long the influence of each spike lasts. The membrane potential $V_m(t)$ represents the neuron's internal state, evolving based on the input signal. It integrates recent spikes and decays toward a resting state when there is no input, with the decay rate controlled by the membrane time constant $\tau_m$. The fire-reset mechanism is captured by the Heaviside step function $H(x)$. This function determines the spiking activity $s(t)$ and triggers the reset of the membrane potential $V_m(t)$ to $V_{reset}$ when $V_m(t)$ exceeds the threshold $\theta$. The following equations describe this behavior:

$$s(t) = \begin{cases} 1 & \text{if } V_m(t) \geq \theta \\ 0 & \text{otherwise} \end{cases} \quad \text{and} \quad V_m(t) = \begin{cases} V_{reset} & \text{after spike} \\ V_m(t) & \text{otherwise} \end{cases}$$

Importantly, the threshold $\theta$ serves as a gating mechanism for spike generation, it's a key factor in the regulation of neural activity [18, 21]. Since the number of spikes directly impacts energy consumption, the threshold offers a natural and lightweight parameter to control energy consumption of a spiking network.

### 2.3 Limitations of Current SNN Design: Fixed-Threshold Training Strategy

In current SNNs, each neuron decides when to fire based on a threshold value $\theta$, which acts as a decision boundary for spike generation. During training, the firing threshold is either initialized and then left unchanged or included as a trainable parameter and updated by the training algorithm. Making thresholds trainable parameters increases the model's flexibility to better fit the data. However, increasing the number of trainable parameters also increases the degrees of freedom of the model, which can make it more complex and harder to regularize. Additionally, having more trainable parameters increases the computational cost of training and can slow convergence, particularly in SNNs where surrogate gradients or time-series processing is involved.

Regardless of the training strategy, having a constant threshold at inference time is the standard approach in most baseline LIF neuron models. While fixed-threshold models often perform well in controlled or well-calibrated settings, they lack the flexibility to adapt their spiking behavior in response to external conditions such as power availability. Moreover, these models are more prone to overfitting and may generalize poorly in noisy or previously unseen environments.

In contrast, biological neurons exhibit a remarkable degree of adaptability, allowing them to operate reliably in noisy and dynamic environments. Studies have shown that this high adaptability arises from multiple factors, including the ability of neurons to regulate its internal parameters, such as firing thresholds and membrane dynamics [25]. For instance, biological neurons maintain stable firing rates by adjusting their firing thresholds, preventing activity from becoming either excessively high or completely silent. By dynamically tuning their internal parameters, biological systems maintain efficient spiking behavior, strong generalization and robust information processing even across fluctuating conditions [17], offering inspiration for more adaptive SNNs designs.

Figure 3 illustrates how threshold adaptation affects the spiking behavior of a neuron model. In the fixed-threshold case (left), the neuron fires regularly because the threshold remains constant, allowing spikes to occur as soon as the membrane potential crosses it. In contrast, the adaptive model (right) increases the threshold over time to control the spiking behavior. As a result, the neuron fires less frequently, since it becomes harder to reach the threshold again. This leads to a sparser pattern of spikes and reflects a mechanism observed in biological neurons. The adaptive behavior can improve energy efficiency and make the model more responsive to meaningful changes in input.

Figure 4 depicts the relationship between classification accuracy and average spike count when a fixed-threshold SNN is evaluated across different thresholds $\theta$. Each point represents the performance of the model under a specific threshold value, with the x-axis showing the average number of spikes per sample and the y-axis showing the corresponding accuracy. The vertical dashed line marks the threshold used during training ($\theta_0 = 1.0$), which results in the highest accuracy. As the threshold increases ($\theta_0 > 1.0$), spiking activity decreases, reducing energy consumption but also significantly lowering accuracy. Conversely, lowering the threshold ($Theta_0 < 1.0$) results in higher spiking rates with only marginal or no gains in robustness. These results indicate that the model is highly sensitive to threshold variations and that careful tuning is required to balance performance and energy efficiency.

## 3 *ASPEN*'S APPROACH

While current SNN implementations demonstrate the potential of neuromorphic computing for ultra-low-power applications, their reliance on fixed intrinsic parameters severely limits their adaptability to dynamic energy constraints. To address this fundamental limitation, CAPUCHIN draws inspiration from biological neural systems where intrinsic plasticity enables neurons to dynamically modulate their excitability in response to environmental conditions.

### 3.1 Bio-Inspired Foundation: *Extrinsic and Intrinsic Plasticity*

Biological neural systems exhibit two fundamental forms of plasticity that have been successfully adapted for SNNs: extrinsic plasticity for synaptic adaptation and intrinsic plasticity for neuronal excitability control [28, 43, 46]. Extrinsic plasticity, or synaptic plasticity, is the ability to adaptive regulate synaptic weights between neurons, which determine the strength of communication among them. This form of plasticity has been the primary focus of most learning algorithms, including biologically inspired approaches such as Spike-Timing Dependent Plasticity [3], as well as more conventional gradient-based methods like surrogate gradient optimization [32]. In contrast, intrinsic plasticity involves modulating internal properties, such as firing thresholds, membrane time constants, or refractory periods. Intrinsic plasticity plays an important role in biological systems by supporting homeostatic regulation and enabling neurons to adapt their excitability based on environmental [20, 33]. Recently, there has been growing interest in using intrinsic plasticity to improve the robustness and learning ability, though it remains less explored in computational models [15, 45].

Figure 5 illustrates the difference between intrinsic and extrinsic plasticity in biological neurons. In intrinsic plasticity, regulating the firing threshold is achieved by modulating ion channels along the axon and soma, such as sodium ($Na^+$) and potassium ($K^+$) channels [33]. By adjusting the density or activity of these channels, neurons can become more or less likely to fire in response to input. In contrast, extrinsic plasticity, involves changes in the strength of connections between neurons. The illustration shows how neurotransmitters released from the pre-synaptic neuron cross the synaptic cleft and bind to receptors on the post-synaptic neuron. Modifications in neurotransmitter release or receptor sensitivity can enhance or weaken synaptic transmission, supporting mechanisms such as long-term potentiation or depression [4, 13].

In the context of SNNs and neuromorphic hardware, adjusting intrinsic parameters like thresholds and time constants offers a practical and efficient way to adapt network behavior. Unlike synaptic weights, which grow rapidly with the number of connections, intrinsic parameters scale with the number of neurons, making them much more scalable. This difference makes intrinsic plasticity especially promising for controlling spiking activity in systems where resources or energy are limited.

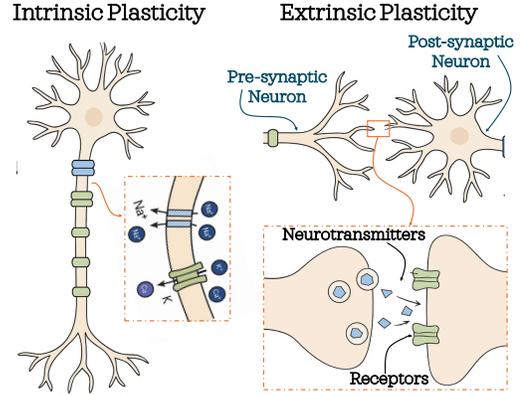

**Figure 5: Intrinsic and Extrinsic Plasticity**

### 3.2 *ASPEN* Framework Overview

Building upon these biological insights, *ASPEN* translates the principles of intrinsic plasticity into a practical neuromorphic computing framework. Rather than relying solely on synaptic weight adjustments, our approach leverages threshold modulation as a lightweight mechanism for real-time energy-accuracy trade-offs. This approach is illustrated in Fig. 6. To mirror this adaptability, *ASPEN* dynamically adjusts the firing thresholds of spiking neurons during inference, enabling real-time control over spiking activity in response to available energy. The system begins with raw sensor data, which is processed through a sensory front-end and converted into spike trains. These are passed into an SNN core, where an energy monitor informs an Energy-Aware Threshold Adaptation module that modulates neuron thresholds based on the current energy budget. This intrinsic parameter tuning regulates spike generation, allowing *ASPEN* to balance energy consumption and classification accuracy. The final spike patterns are decoded to classify physical activities such as running, cycling, walking, and swimming. This threshold modulation serves as an internal energy-aware mechanism that adapts SNN behavior to varying energy contexts without retraining or structural changes.

At its core, *ASPEN* implements a compact SNN tailored for real-time inertial data processing. The architecture consists of an input layer followed by three fully connected hidden layers with 48 Leaky Integrate-and-Fire (LIF) neurons each, and a final output layer for activity classification. The input is a spike-encoded representation of IMU signals, and the network is trained using supervised learning with backpropagation-through-time (BPTT) and surrogate gradients. Training uses cross-entropy loss and the Adam optimizer with a standard learning rate schedule. To improve generalization and energy adaptability, *ASPEN* introduces stochastic variability in firing thresholds during training by sampling from a uniform distribution. This regularization encourages the network to form robust internal representations that are resilient across a spectrum of spiking behaviors.

At inference time, *ASPEN* supports two complementary energy-aware adaptation strategies. In single-model adaptation, a network trained with stochastic thresholds adjusts its internal thresholds post-deployment to match energy constraints. In model-switching adaptation, *ASPEN* selects among multiple pre-trained models, each optimized for a specific spiking regime. Lower thresholds lead to denser spiking and higher task accuracy, while higher thresholds reduce activity and power consumption. This dual strategy allows *ASPEN* to scale its computational cost in fine-grained or coarse-grained ways, supporting deployment in ultra-low-power scenarios, including always-on or battery-free wearable devices. The next section details the design of *ASPEN*, its technical challenges, and innovations.

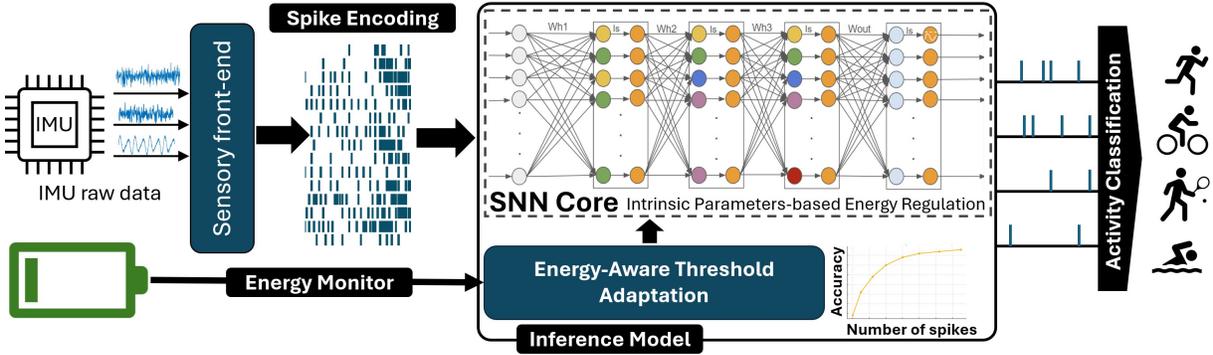

**Figure 6: Overview of *ASPEN*'s design**

## 4 ADAPTIVE ENERGY-AWARE DESIGN

In always-on wearables, it is critical for spiking neural networks to adapt their spiking activity in response to changing resource availability, such as power fluctuations or shifting computational budgets, without requiring retraining or structural modification. This adaptability allows the system to maintain reliable performance across a range of operating conditions. To support this goal, we present two complementary strategies that enable flexible, energy-aware behavior in SNNs. The first strategy introduces stochastic variability into the neuron thresholds during training, allowing the network to develop internal representations that are robust to different spiking conditions induced by threshold variations. The second strategy leverages the robustness gained through threshold stochasticity to enable long-range adaptation at inference time by switching between models configured for different operating conditions, such as accuracy or energy consumption.

### 4.1 Strategy #1: Training with Stochastic Thresholds

We propose a training approach that improves SNN robustness and efficiency by introducing stochasticity into threshold parameters during training. In conventional fixed-threshold training, neurons become highly sensitive to specific threshold values, resulting in significant degradation of the network if the threshold is modified during deployment. This limits the ability of the network to adapt to energy constrains in real-world scenarios where operating conditions can change. So, instead of using a fixed threshold (for example, $\theta_0 = 1.0$), we introduce variability by sampling threshold values from a distribution during training. This can be done in two ways: (1) sampling from a *continuous uniform distribution*, such as $\theta \sim \mathcal{U}(\theta_{\min}, \theta_{\max})$, or (2) sampling from a *discrete set of threshold values*, for example $\theta \in \{\theta_1, \theta_2, \ldots, \theta_n\}$. The discrete approach allows thresholds to be chosen from a fixed, predefined set and is particularly relevant for digital neuromorphic hardware with a limited number of threshold levels supported. Both methods expose the network to a diverse range of excitability conditions during training, which encourages the development of robust internal representations. Rather than relying on narrowly tuned responses that perform well only under fixed, ideal thresholds, the network learns to generate stronger and more consistent activations that remain effective across a wider range of threshold values.

This can be understood by analyzing the expected spike probability. In the traditional fix-threshold LIF neuron, a spike occurs when the membrane potential $V_m(t)$ exceeds a fix threshold $\theta_0$, such that the probability of spiking at time $t$ is: $\mathbb{P}(s(t) = 1) = \mathbb{P}(V_m(t) \geq \theta_0)$. Assumming $V_m(t)$ follows a probability density function $p_{V_m}(v)$, the probability becomes

$$\mathbb{P}(s(t) = 1) = \int_{\theta_0}^{\infty} p_{V_m}(v)dv$$

Since $s(t) \in \{0, 1\}$, then the expected probability of a spike at time $t$ becomes

$$\mathbb{E}[s(t)] = \mathbb{P}(V_m(t) \geq \theta_0) = \int_{\theta_0}^{\infty} p_{V_m}(v)dv$$

From this equation, we observe that increasing the threshold reduces the probability of a spike at time $t$, while decreasing the threshold increases it. Thus, the threshold $\theta_0$ can be used as a control parameter for modulating the spiking activity of the neuron.

In our approach, we replace the fixed threshold with a stochastic threshold $\theta$ sampled from a uniform distribution. Because the threshold is now a random variable, the spike probability also becomes a random variable $P(V_m(t) \geq \theta)$. To compute the expected spike probability under this stochastic threshold, we average over all possible threshold values that the threshold can take.

■ **Continuous threshold sampling.** In the case of a continuous threshold distribution, where $\theta \sim \mathcal{U}(\theta_{\min}, \theta_{\max})$, the expected spike probability is given by

$$\mathbb{E}[s(t)] = \frac{1}{\theta_{\max} - \theta_{\min}} \cdot \int_{\theta_{\min}}^{\theta_{\max}} \left[ \int_{\theta}^{\infty} p_{V_m}(v) \, dv \right] d\theta$$

This expression quantifies how often the neuron is expected to spike on average when the threshold varies continuously across a range. The inner integral calculates the probability that the membrane potential $V_m(t)$ exceeds a specific threshold $\theta$, corresponding to the likelihood of spiking under that condition. The outer integral then averages this probability over the entire range of possible thresholds. Lower thresholds increase the chance of spiking, while higher thresholds reduce it. The result is an average measure of spiking behavior across the full threshold distribution.

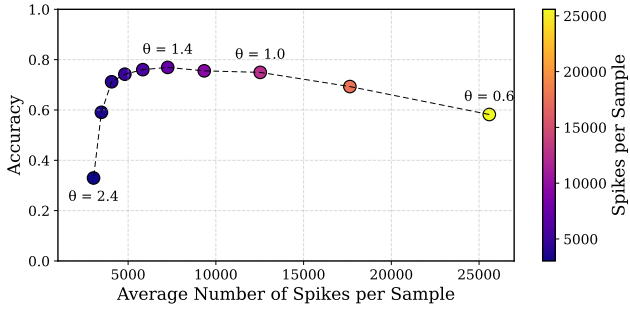

**Figure 7: Sensitivity Analysis Across Threshold Space in a Stochastic-Threshold Trained Model**

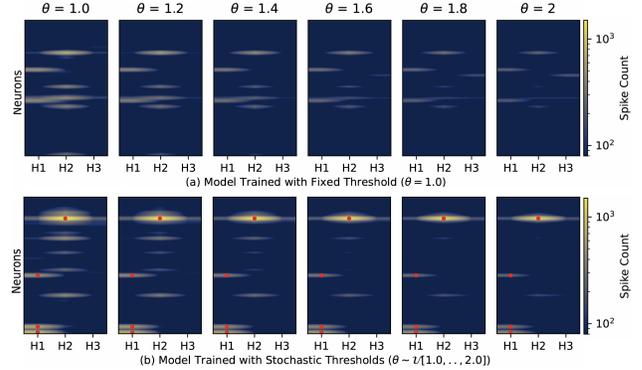

**Figure 8: Spiking activity across hidden layers under varying thresholds. Each subplot shows the average spike count per neuron for a given threshold, with hidden layers H1, H2, and H3 along the x-axis and neuron indices on the y-axis.**

■ **Discrete threshold sampling.** Alternatively, in discrete threshold sampling, the threshold is drawn from a finite set of values, $\Theta = \{\theta_1, \theta_2, \ldots, \theta_n\}$. Using discrete threshold sampling during training offers a computationally efficient approximation of continuous threshold variability, enabling the network to experience sufficient threshold diversity without incurring the full overhead of continuous sampling. In this case, the expected spike probability becomes a finite sum:

$$\mathbb{E}[s(t)] = \frac{1}{|\Theta|} \sum_{\theta \in \Theta} \int_{\theta}^{\infty} p_{V_m}(v) \, dv$$

Here, $|\Theta|$ is the number of discrete threshold values. This formulation serves the same purpose as the continuous case: it averages the spiking probability over multiple excitability levels. Both formulations encourage the network to learn internal representations that are robust to variations in neuron firing thresholds.

To compare the behavior of the expected spike probability under stochastic threshold with that under a fixed threshold, we can define the probability function $f(\theta) = P(V_m(t) \geq \theta)$, which represents the probability of a spike as a function of the threshold. Since the probability decreases when the threshold $\theta$ increases, $f(\theta)$ is a monotonically decreasing function.

Because $f(\theta)$ is both, monotonically decreasing and, in most practical cases (e.g., when $V_m(t)$ is normally distributed) concave, we can apply Jensen's inequality. Jensen's inequality states that for a concave function $f$ and a random variable $\theta$

$$f(\mathbb{E}[\theta]) > \mathbb{E}[f(\theta)]$$

This implies that the spike probability using a fixed threshold set to the average of the stochastic thresholds distribution (e.g. $\theta_0 = \mathbb{E}[\theta]$) is greater than the average spike probability over the distribution of thresholds. Hence, when the fixed threshold $\theta_0$ is less than the mean of the stochastic threshold distribution, the stochastic approach results in lower expected spiking activity than using $\theta_0$.

■ **Advantage of stochastic thresholds.** Two key benefits in training with stochastic thresholds: First, it enables a continuous and graceful trade-off between accuracy and spiking activity because the network is exposed to a wide range of threshold values during training (see Fig 7). It learns to generalize across different spiking regimes, instead of overfitting to a single threshold. It also allows post-deployment adjustments to the threshold to modulate the spiking rate gradually, maintaining usable performance over a range of energy conditions instead of showing a sudden drop in accuracy.

Second, the use of stochastic thresholds during training acts as an effective form of regularization. When the neuron excitability varies during learning, the network cannot rely on fine-tuned threshold-specific patterns. Instead, it learns to respond only to the most reliable and informative inputs (See Fig 8). From the perspective of the probabilistic model, this corresponds to lowering the expectid spike probability $\mathbb{E}[s(t)]$ as shown in our modeling analysis. By sampling thresholds from a distribution during training, the network effectively suppresses weak or marginal activations that would only trigger spikes under low thresholds. This leads to sparser and more efficient spiking activity, as the network requires fewer spikes to reach a given level of accuracy. As a result, models trained with this approach often achieve better energy efficiency compared to those trained with fixed thresholds.

Learning under variable thresholds allows the trained network to exploit threshold modulation at inference time. This capability is particularly helpful for energy-constrained or adaptive neuromorphic systems, where thresholds may be adjusted on-the-fly to control spiking activity and power consumption. Importantly, because the model has already learned to operate under threshold variability, no retraining or structural modification is needed during deployment. This enables runtime energy adaptation with minimal accuracy loss, supporting efficient and flexible inference in environments with dynamic power budgets, i.e., always-on settings.

### 4.2 Strategy #2: Long-Range Adaptation with Threshold Stochasticity & Model Switching

We introduce a model-switching concept as a complementary strategy to extend the dynamic range of energy-accuracy trade-offs. Model switching is a strategy in which a system dynamically selects between multiple pre-trained models, each optimized for a specific operating condition, such as accuracy, latency, or energy consumption [6, 22]. Rather than relying on a single network to perform well across all scenarios, model switching leverages a set of specialized networks and transitions between them based on runtime constraints or environmental feedback. This is similar to low-power deep learning methods for mobile and embedded applications, where switching between compressed or early-exit models helps manage energy and performance trade-offs [42, 44].

While threshold modulation enables continuous and graceful control within a single model, model switching allows the system to transition between multiple pre-trained models, each offering a better accuracy–spike trade-off in a specific energy regime. The main advantage of using a single model is that controlling spiking activity can be achieved by simply adjusting the neuron thresholds at inference time. This makes threshold modulation a lightweight and efficient strategy for real-time energy control, as it avoids the overhead of loading or reconfiguring a new model. However, a single model is typically limited in its ability to deliver optimal performance across all energy conditions. Threshold tuning alone cannot always compensate for the limitations of the model, especially at the extremes of the accuracy–efficiency spectrum. In contrast, switching between models that present better accuracy-spikes tradeoff for different operating points extends the overall adaptation range, at the cost of a more complex runtime system.

To enable scalable adaptation across a wide range of energy constraints, we structure our system around the Pareto front defined by the trade-off between classification accuracy and spiking activity. Each segment on this front corresponds to a model optimized for a specific balance between performance and energy efficiency. By configuring our system in this way, we achieve two levels of adaptation: *a coarse-grained control via model switching*, and *a fine-grained control via threshold modulation*. The combination of these two levels yields a flexible and efficient inference system capable of maintaining near-optimal performance across a broad spectrum of operating conditions.

Our model switching strategy operates based on available energy budget at inference time. For this, we first construct the Pareto front offline, by training multiple models that represent optimal trade-offs between classification accuracy and spiking activity. We denote this collection of models as: $\mathcal{M} = \{M_1, M_2, \ldots, M_n\}$. For each model $M_i$, we identify $K_i$ distinct optimal operating segments over the threshold space where the model offers best trade-off between accuracy vs. spiking activity. We denote these segments as

$$\Theta_i^* = \left\{ \Theta_i^{(1)}, \Theta_i^{(2)}, \ldots, \Theta_i^{(K_i)} \right\}$$

where each segment $\Theta_i^{(k)} \subseteq \Theta_i$ corresponds to a locally optimal trade-off region. So, the set of locally optimal operating points is:

$$\mathcal{R}_i^* = \bigcup_{k=1}^{K_i} \left\{ (A_i(\theta), S_i(\theta)) \,\middle|\, \theta \in \Theta_i^{(k)} \right\}$$

where $A_i(\theta)$ represents the accuracy and $S_i(\theta)$ the spiking activity. It is important to note that a single model $M_i$ can exhibit multiple locally optimal segments within its operating range. This occurs when different threshold values lead to distinct, favorable trade-offs, allowing the same model to contribute on multiple segments to the overall Pareto front. The global Pareto front is defined then as

$$\mathcal{F} \subseteq \bigcup_i \mathcal{R}_i^*$$

with each $\mathcal{R}_i^*$ representing the piecewise Pareto-optimal behavior of model $M_i$. This approach avoids the need for online evaluation to select the appropriate model, reducing computational overhead while still enabling continuous and graceful transitions between operating points. It also ensures that the selected configuration is well-aligned with the available resources to enable efficient and scalable deployment across a wide range of operating environments. Hence, the system can gracefully degrade or enhance its performance based on energy availability, so it's well-suited for real-world deployment in energy-harvesting or always-on applications.

### 4.3 Adaptive Strategy During Inference-Time

At inference time, controlling energy consumption is critical for deploying spiking neural networks in resource-constrained environments. Our framework supports two complementary strategies for energy-aware inference, both relying on threshold modulation. In the single-model setting, the network is trained with stochastic threshold variability, which enables it to generalize across a wide range of firing thresholds. This allows the system to adjust the threshold dynamically during inference based on the available energy budget. As a result, it can reduce spiking activity while maintaining a gradual and graceful degradation in accuracy. In the multi-model setting, we extend this approach by combining several pre-trained models, each designed to operate efficiently within a specific region of the accuracy and spike trade-off space. During inference, the system selects the model that best matches the current energy constraint, and then further refines performance by adjusting the threshold within that model. This combination of model selection and threshold tuning provides flexible and efficient inference across a wide range of operating conditions.

In the single-model setting, the model trained with stochastic thresholds defines a continuous set of operating points that are parameterized by the threshold value $\theta$. This set is given by

$$\mathcal{R}_i = \{ (A_i(\theta), S_i(\theta)) \,|\, \theta \in \Theta_i \}$$

where $A_i(\theta)$ and $S_i(\theta)$ represent the accuracy and spiking activity of the model at threshold $\theta$, respectively. At inference time, the system selects a threshold value $\theta^* \in \Theta_i$ to control the trade-off between energy consumption and performance. The goal is to satisfy the energy constraint $E$ while maximizing accuracy, which can be formulated as

$$\theta^* = \arg\max_{\theta \in \Theta_i} A_i(\theta) \quad \text{subject to } S_i(\theta) \leq E$$

In the multi-model setting, the system evaluates the current energy constraints and selects the most appropriate model and operational range from the Pareto front. To do this, we define a selection function that maps the energy budget $E$ to a optimal operation range and threshold

$$(\mathcal{R}_i^*, \theta^*) = \text{Select}(E), \text{ such that}$$

$$\theta^* = \arg\max_{\theta \in \Theta_i} A_i(\theta) \quad \text{subject to } (A_i(\theta), S_i(\theta)) \in \mathcal{R}_i^*, \ S_i(\theta) \leq E$$

This two-level inference strategy provides both coarse and fine-grained control. Model switching enables broad adaptation across energy regimes by transitioning between models optimized for different accuracy–efficiency trade-offs. Threshold adaptation within each selected model further refines behavior to match real-time resource constraints. Together, these mechanisms allow neuromorphic systems to adapt efficiently to dynamic energy budgets while preserving efficient and accurate operation.

## 5 IMPLEMENTATION

We implemented all models using the Rockpool framework [31], which provides native support for training and evaluating spiking neural networks and is integrated with the SynSense Xylo HDK hardware platform. This integration allowed us to directly test our system on the Xylo neuromorphic chip, ensuring compatibility and efficient deployment on real-world spiking hardware. Training and evaluation were performed on an NVIDIA A100 GPU.

Our architecture is based on the SynNet architecture [7, 8] and consists of a fully connected SNN with an input layer, three hidden layers of LIF neurons, and an output layer. Each hidden layer contains 48 spiking neurons, chosen to balance representational capacity with low computational overhead. The input layer comprises 15 neurons, corresponding to the 15 output channels of our spike-based IMU encoder. The output layer contains a variable number of neurons depending on the classification task (e.g., 3 or 6 classes).

To enhance temporal processing capabilities, SynNet employs a pyramidal configuration of synaptic time constants across neurons in each hidden layer. Instead of using a single synaptic time constant $\tau_s$ for all neurons, each layer includes multiple fixed synaptic time constants defined as $\tau_{s_n} = 2^n \times 1$ ms. Neurons are evenly divided among these values, ensuring a balanced representation of short-, mid-, and long-term temporal integration. For example, layers with two synaptic time constants assign half of the neurons a constant of $\tau_{s_1} = 2$ ms and the other half $\tau_{s_2} = 4$ ms. In layers with four synaptic time constants, neurons are evenly distributed across $\tau_{s_1} = 2$ ms, $\tau_{s_2} = 4$ ms, $\tau_{s_3} = 8$ ms, and $\tau_{s_4} = 16$ ms. The system adopts the configuration $\tau = [2, 4, 8]$, meaning the first hidden layer uses two synaptic time constants, the 2nd uses 4, and the 3rd uses 8.

The membrane time constant $\tau_m$ is kept fixed across all neurons to maintain stable internal dynamics and $\tau_m = 2ms$ . This configuration not only improves temporal generalization but is also compatible with the Xylo HDK hardware, which supports fixed parameter neuron models optimized for low-power execution. The raw IMU signals were encoded into spike trains using the encoder provided by the Rockpool framework. This encoder replicates the functionality of the hardware front-end implemented in the SynSense Xylo platform, ensuring full compatibility between our simulations and neuromorphic deployment. By using the same encoding strategy in both software and hardware, we maintain consistency in the input representation and enable a direct transition from model development to real-world execution on the Xylo HDK.

The spike encoder consists of a filter bank, rectifiers, and IAF neurons. The filter bank decomposes each IMU input channel (X, Y, Z) into different frequency components. For each channel, five bandpass filters are applied with cutoff frequencies of 1–2Hz, 2–4Hz, 4–8Hz, 8–16Hz, and 16–32Hz, respectively. The outputs of these filters are then rectified and used as input signals for IAF neurons. Each IAF neuron integrates its input over time as membrane voltage, emits a spike when the voltage crosses a threshold, and then resets its membrane potential using a subtraction-based reset mechanism. In total, the spiking encoder receives three input channels (X, Y, Z) and produces fifteen output spike trains, corresponding to one IAF neuron per frequency band per channel.

To encourage robustness to threshold variability, we introduced stochasticity during training by randomly sampling the neuron firing threshold from a uniform distribution. We trained multiple models using both continuous and discrete uniform distributions, each spanning different threshold ranges. Specifically, we explored the following configurations:

- **Fixed Threshold:** $\theta = 1.0$
- **Continuous Uniform Sampling:** $\theta \sim \mathcal{U}(1.0, 1.5)$; $\theta \sim \mathcal{U}(1.0, 2.0)$; $\theta \sim \mathcal{U}(1.0, 3.0)$
- **Discrete Uniform Sampling:** $\theta \sim \mathcal{U}(\{1.0, \ldots, 1.5\})$; $\theta \sim \mathcal{U}(\{1.0, \ldots, 2.0\})$; $\theta \sim \mathcal{U}(\{1.0, \ldots, 3.0\})$

For each training batch, a new threshold value was sampled and applied uniformly across all spiking neurons in the hidden layers. Baseline models were implemented using fixed-threshold training with identical architectures.

For training, we used the Adam optimizer with an initial learning rate of $1 \times 10^{-3}$ or $1 \times 10^{-4}$, depending on the dataset. Training employed ReLU-based surrogate gradients and a cross-entropy loss function. Models were trained for 500 to 1000 epochs, with the exact number of epochs determined based on dataset characteristics and convergence behavior. For some models, we incorporated a cosine annealing learning rate scheduler with warm restarts. This scheduler stabilizes training by gradually decreasing the learning rate and periodically resetting it, helping the model escape local minima and improving overall convergence.

To analyze the generalization capabilities of the models under threshold modulation, we evaluated each trained model on the test set by systematically varying the firing threshold across a predefined range. Specifically, we swept the threshold values from 0.6 to 2.4 in fixed increments of 0.2. For each threshold in this range, the entire test set was passed through the network without retraining or modifying any other parameters. This procedure allowed us to observe how the model performed in terms of classification accuracy and spiking activity as a function of the firing threshold. By analyzing these results, we quantified the robustness of the model to changes in spiking conditions and its ability to gracefully adapt its energy–performance trade-off at inference time.

## 6 EVALUATION

In this section, we present a comprehensive evaluation of *ASPEN*'s adaptive threshold approach across three key dimensions. First, we demonstrate *ASPEN*'s superior robustness compared to fixed-threshold baselines through systematic threshold variation analysis, showing how stochastic training enables graceful degradation under energy constraints. Second, we extend this analysis to construct Pareto-optimal operating curves by strategically combining multiple threshold sampling strategies, revealing how different approaches excel in specific energy regimes. Finally, we validate the generalizability of our findings through cross-dataset and neuromorphic hardware evaluation, confirming that *ASPEN*'s benefits extend across diverse settings.

### 6.1 Experimental Setup

To investigate the effect of stochastic thresholds during training on model robustness and energy adaptability, we design a comparative study between fixed-threshold and stochastic-threshold training approaches. We focus on evaluating how threshold variability during training affects a model's ability to maintain performance across different energy operating points at inference time.

■ **Training Configurations.** We establish two primary training configurations for comparison:

- **Baseline (Fixed-threshold)**: Models trained with a constant threshold $\theta = 1.0$ throughout training
- **ASPEN (Stochastic-threshold)**: Models trained with thresholds sampled from a defined range (e.g. $\mathcal{U}(1.0, 1.5)$) for each training batch

As described in §4.1, the stochastic approach introduces controlled variability in neuronal activation during training in order to promote robustness to threshold changes during deployment. Both configurations use identical network architectures and training procedures, differing only in threshold sampling strategy.

■ **Robustness Evaluation Protocol.** Post-training, we systematically evaluate each model's robustness by testing across a range of fixed inference thresholds from $\theta = 0.6$ to $\theta = 2.4$ in increments of 0.2. This evaluation protocol simulates varying energy constraints that might occur during real-world deployment, where threshold adjustment serves as a mechanism for energy-accuracy trade-offs. For each threshold value, we measure classification accuracy and average spike count across the complete test set without any parameter updates.

■ **Evaluation Metrics.** To quantitatively assess the energy-accuracy trade-offs enabled by different training strategies, we define two complementary metrics:

- **Relative Accuracy Reduction** ($\Delta\mathrm{Acc}_{rel}$) measures the percentage change in classification accuracy relative to the fixed-threshold baseline:

$$\Delta\mathrm{Acc}_{rel} = \frac{\mathrm{Acc}_{model} - \mathrm{Acc}_{baseline}}{\mathrm{Acc}_{baseline}}$$

where negative values indicate accuracy loss and positive values indicate accuracy gains.

- **Relative Spike Reduction** ($\Delta\mathrm{Spk}_{rel}$) quantifies the percentage change in spikes count relative to the baseline:

$$\Delta\mathrm{Spk}_{rel} = \frac{\mathrm{Spk}_{model} - \mathrm{Spk}_{baseline}}{\mathrm{Spk}_{baseline}}$$

where negative values indicate a reduction in spike count, and positive values indicate an increase.

## 6.2 Stochastic vs. Fixed Threshold Training

Fig. 9 compares the accuracy-spike trade-offs between models trained with stochastic thresholds vs. fixed thresholds, evaluated on the UCI-HAR dataset [37]. During evaluation, thresholds were systematically adjusted from 1-2 in increments of 0.2, with resulting accuracy and spike counts recorded for each operating point. *ASPEN* provides a clear superset of capabilities compared to fixed threshold training. Our stochastic threshold model achieves the same (or a bit higher) maximum accuracy-spike performance as the fixed-threshold baseline, but demonstrates dramatically improved adaptability across varying energy constraints.

■ **Graceful Degradation Under Energy Constraints.** The stochastic model exhibits graceful degradation characteristics essential for duty-cycled operation in low-power devices as well as enhanced robustness to variations in real-world operating conditions. Fig. 9 shows that *ASPEN* maintains ~2% accuracy loss across an 8,000-spike range (~5,000 to 13,000 spikes) and ~4% loss across an 11,000-spike range (~4,000 to 15,000 spikes). This broad operating envelope enables real-time adaptation to fluctuating energy budgets without catastrophic performance drops.

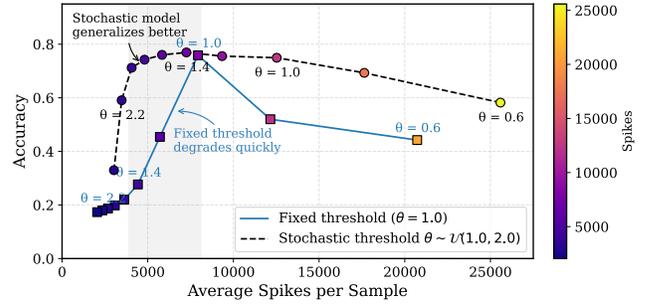

**Figure 9: Sensitivity Analysis Across Threshold Space in a Stochastic-Threshold Trained Model.**

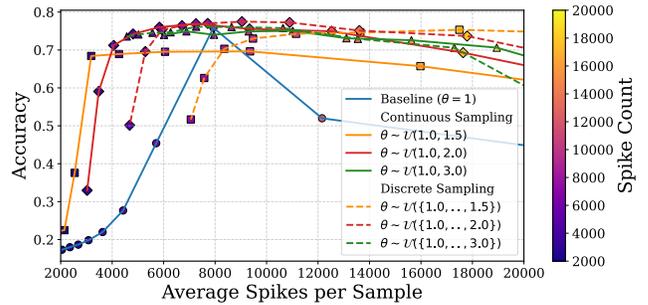

**Figure 10: Accuracy vs. Average Spike Count Across Threshold Sampling Strategies.**

In contrast, the fixed-threshold model is more brittle: ~2% accuracy loss occurs within only a 300-spike window (~7,900 to 8,200 spikes), and ~4% loss spans just 800 spikes (~7,800 to 8,600), making the system vulnerable to threshold miscalibration.

■ **Practical Implications for Wearable Systems.** The wide operating range directly enables duty-cycled sensor operation, where systems must dynamically adjust computational load based on available energy. For example, during low-energy periods, the system can increase thresholds to operate at 5,000 spikes with only 2% accuracy loss, while during high-energy periods, it can operate at 8,000 spikes for maximum performance. Importantly, when domain shifts or unexpected operating conditions cause threshold miscalibration, *ASPEN*'s graceful degradation prevents system failure. The fixed-threshold approach risks complete functionality loss outside its narrow operating window, making it unsuitable for real-world deployment where operating conditions vary.

■ **Energy Efficiency Demonstration.** In the illustrated example, *ASPEN* achieves a 1.95× reduction in spike count with only 2.2% accuracy loss, demonstrating favorable energy-accuracy trade-offs. This efficiency gain enables extended battery life in always-on wearable applications while maintaining acceptable performance levels. The analysis reveals that while *ASPEN* demonstrates robustness across a wide operating window, beyond $\theta > 1.8$ accuracy declines rapidly as the network's capacity to process information under extremely sparse spiking conditions becomes limited. This defines the practical lower bound for energy-constrained operation.

Overall, **ASPEN**'s stochastic threshold training provides a **10-40× wider robust operating range** compared to fixed-threshold approaches, enabling practical deployment in energy-variable environments while maintaining the same peak performance capabilities. This robustness is critical for wearable neuromorphic systems that must adapt to changing energy availability without compromising essential functionality.

### 6.3 Pareto-Optimal Energy-Accuracy Trade-offs

Our previous results showed that **ASPEN** with a single stochastic threshold sampling strategy ($\mathcal{U}(1.0, 1.5)$) significantly outperforms fixed-threshold baselines in terms of robustness and energy efficiency. This raises a critical question: *can we leverage multiple sampling strategies to create a family of models that extends the dynamic range even further, for greater robustness and efficiency across an even wider operational range than any single model?*

To answer this question, we systematically evaluate how different threshold sampling strategies affect the energy-accuracy trade-off, then demonstrate how combining these strategies creates a comprehensive Pareto front that maximizes adaptability across diverse energy constraints.

■ **Impact of Sampling Strategy on Model Performance.** Fig. 10 illustrates how threshold sampling intervals affect the spike-accuracy trade-off at inference time. We compare a baseline model trained with fixed threshold $\theta = 1.0$ against several models trained with stochastic thresholds sampled from uniform distributions with varying ranges: $\mathcal{U}(1.0, 1.5)$, $\mathcal{U}(1.0, 2.0)$, and $\mathcal{U}(1.0, 3.0)$. Each curve represents a different sampling strategy, with dashed lines indicating discrete threshold sampling variants.

*Moderate Stochasticity Optimizes Robustness.* Models trained with moderate threshold stochasticity, such as $\mathcal{U}(1.0, 1.5)$, demonstrate robust accuracy across broader threshold and spike budget ranges. This mild stochasticity acts as an effective regularizer, improving generalization to different threshold regimes while maintaining peak performance. Importantly, these models enable *post-deployment threshold adjustment* for energy-accuracy trade-offs without network retraining which is a critical capability for always-on systems.

*Discrete vs. Continuous Sampling.* Discrete threshold sampling consistently outperforms continuous sampling across multiple ranges. For $\mathcal{U}(1.0, 1.5)$, discrete sampling achieves 1.9× spike reduction with only 0.1% accuracy loss compared to fixed-threshold baselines. For $\mathcal{U}(1.0, 2.0)$, discrete sampling delivers 3.3× spike reduction with 1.2% accuracy loss. This superior performance suggests discrete sampling acts as a stronger regularizer, promoting stable neuron activation while reducing hardware storage requirements.

*Excessive Variability Degrades Performance.* Wide threshold distributions like $\mathcal{U}(1.0, 3.0)$ result in lower peak accuracy, likely due to unstable neuron behavior during training. While these models achieve lower spike counts, the accuracy penalty makes them unsuitable for performance-critical applications.

■ **Constructing the Pareto Front for Maximum Adaptability.** Figure 11 presents the key insight: *different sampling strategies excel in different regions of the energy-accuracy space.* When deployment constraints limit the system to a single model, the discrete mid-range model $\mathcal{U}(1.0, 2.0)$ represents the optimal choice, delivering the highest accuracy (over 85% accuracy) and exceptional energy efficiency (only 5,000-6,000 spikes per sample). However, to achieve the best performance across all energy regimes, we can strategically connect models from different sampling strategies to construct a comprehensive Pareto front that maximizes efficiency across the entire operational range.

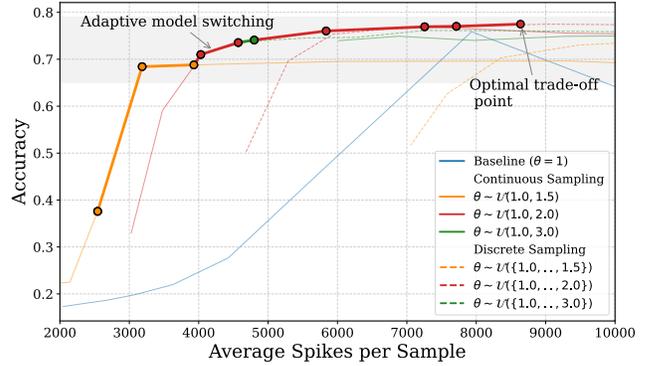

**Figure 11: Pareto Front for Long-Range Adaptation Across Multi-Stochastic Threshold Models.**

- **Ultra-low power regime**: Wider distributions provide acceptable accuracy at extremely low spike counts.
- **Balanced regime**: Moderate distributions ($\mathcal{U}(1.0, 1.5)$, $\mathcal{U}(1.0, 2.0)$) offer optimal accuracy-efficiency trade-offs.
- **Performance-critical regime**: Conservative distributions maintain high accuracy when energy is less constrained.

These results underscore that discrete threshold training not only improves energy efficiency and hardware compatibility, but also enables flexible, inference-time control, making it particularly suited for wearable, edge, and batteryless neuromorphic applications.

### 6.4 Cross-Dataset Validation

To validate the generalizability of **ASPEN**'s adaptive threshold approach, we evaluate our method across three publicly available human activity recognition datasets: KU-HAR [40], UCI-HAR [37], and UniMiB-SHAR ADL [30]. Each dataset presents unique challenges in terms of sensor modalities, sampling rates, activity classes, and noise characteristics. Testing under diverse conditions reveals the system's classification performance and its ability to adapt to varying data distributions and energy constraints.

Table 1 presents comprehensive results across different threshold ranges for all sampling strategies, revealing consistent patterns that validate **ASPEN**'s robustness across diverse data characteristics and task complexities.

■ **Superiority of Stochastic Approaches.** Across all datasets, **ASPEN** models consistently outperform fixed-threshold baselines when evaluated away from the training threshold. While fixed-threshold models achieve peak performance only at their trained threshold ($\theta = 1.0$), they suffer catastrophic degradation at higher thresholds. For example, on KU-HAR, the fixed-threshold model drops from 84.7% to just 20.2% accuracy when $\theta$ increases from 1.0 to 1.4, whereas **ASPEN** models has a 2-3% accuracy reduction in the same regime.

■ **Discrete Sampling Delivers Superior Efficiency.** The bolded entries in Table 1 highlight optimal accuracy-spike operating points for each threshold regime. Discrete sampling strategies consistently achieve these optimal points, for example $\mathcal{U}(\{1.0, .., 1.5\})$ excels in moderate threshold regimes ($\theta = 1.2$), achieving 84.8% accuracy with only 8,570 spikes on KU-HAR whereas $\mathcal{U}(\{1.0, .., 2.0\})$ dominates higher threshold regimes ($\theta = 1.6$), maintaining 83.7% accuracy with just 4,776 spikes.

Table 1: Accuracy and spike count across thresholds for different sampling methods and datasets.

| Dataset | Type | Sampling | $\theta = 1.0$ | | $\theta = 1.2$ | | $\theta = 1.4$ | | $\theta = 1.6$ | | $\theta = 1.8$ | | $\theta = 2.0$ | |
|---|---|---|---|---|---|---|---|---|---|---|---|---|---|---|
| | | | $\Delta Acc_{rel}$ | $\Delta Spk_{rel}$ | $\Delta Acc_{rel}$ | $\Delta Spk_{rel}$ | $\Delta Acc_{rel}$ | $\Delta Spk_{rel}$ | $\Delta Acc_{rel}$ | $\Delta Spk_{rel}$ | $\Delta Acc_{rel}$ | $\Delta Spk_{rel}$ | $\Delta Acc_{rel}$ | $\Delta Spk_{rel}$ |
| **KU-HAR [40]** | Base. | $\theta = 1.0$ | 84.7% | 16017 | | | | | | | | | | |
| | Cont. | $\theta \sim \mathcal{U}(1.0, 1.5)$ | -4.81% | -10.0% | -0.98% | -30.8% | -3.88% | -44.7% | -12.55% | -55.5% | -33.86% | -63.0% | -68.13% | -69.0% |
| | Cont. | $\theta \sim \mathcal{U}(1.0, 2.0)$ | -13.34% | +37.0% | -7.19% | +10.3% | -6.10% | -7.8% | -5.50% | -20.5% | -7.90% | -29.6% | -8.76% | -37.0% |
| | Cont. | $\theta \sim \mathcal{U}(1.0, 3.0)$ | -22.26% | +37.9% | -14.92% | -2.4% | -13.15% | -27.6% | -13.03% | -44.8% | -14.46% | -56.7% | -13.26% | -65.5% |
| | Disc. | $\theta \sim \mathcal{U}(\{1.0, .., 1.5\})$ | -1.72% | -27.4% | +0.12% | -46.5% | -3.53% | -58.7% | -11.52% | -67.3% | -25.70% | -73.1% | -47.41% | -77.0% |
| | Disc. | $\theta \sim \mathcal{U}(\{1.0, .., 2.0\})$ | -9.19% | -33.5% | -4.65% | -51.2% | -1.92% | -62.5% | -1.12% | -70.2% | -2.36% | -75.5% | -8.10% | -80.0% |
| | Disc. | $\theta \sim \mathcal{U}(\{1.0, .., 3.0\})$ | -20.35% | +107.0% | -15.81% | +55.6% | -12.86% | +18.1% | -10.59% | -10.0% | -10.68% | -30.4% | -6.81% | -44.3% |
| **UCI-HAR [37]** | Base. | $\theta = 1.0$ | 75.86% | 7933 | | | | | | | | | | |
| | Cont. | $\theta \sim \mathcal{U}(1.0, 1.5)$ | -8.21% | +18.0% | -8.34% | -23.7% | -9.14% | -46.2% | -9.81% | -59.9% | -50.45% | -67.9% | -70.37% | -73.0% |
| | Cont. | $\theta \sim \mathcal{U}(1.0, 2.0)$ | -1.21% | +57.9% | -0.41% | +17.7% | +1.40% | -8.5% | +0.21% | -26.4% | -2.16% | -39.3% | -6.13% | -48.8% |
| | Cont. | $\theta \sim \mathcal{U}(1.0, 3.0)$ | -6.92% | +138.8% | -4.32% | +97.0% | -3.65% | +65.2% | -1.07% | -59.9% | -1.28% | +17.3% | -2.44% | +0.3% |
| | Disc. | $\theta \sim \mathcal{U}(\{1.0, .., 1.5\})$ | -2.16% | +194.2% | -0.70% | +120.5% | -1.69% | 72.0% | -2.10% | +40.5% | -3.76% | +19.4% | -7.32% | 5.4% |
| | Disc. | $\theta \sim \mathcal{U}(\{1.0, .., 2.0\})$ | -0.98% | +71.5% | +1.85% | +37.4% | +2.14% | 14.0% | +1.52% | -2.8% | +0.74% | -15.4% | -0.67% | -25.2% |
| | Disc. | $\theta \sim \mathcal{U}(\{1.0, .., 3.0\})$ | -3.85% | +70.8% | -0.12% | +34.1% | +0.26% | +8.9% | +0.37% | -8.4% | -1.58% | -20.9% | -1.74% | -29.8% |
| **UniMiB ADL [30]** | Base. | $\theta = 1.0$ | 88.06% | 5734 | | | | | | | | | | |
| | Cont. | $\theta \sim \mathcal{U}(1.0, 1.5)$ | -2.73% | +59.71% | -0.53% | -0.91% | -4.78% | -32.80% | -15.88% | -51.62% | -43.23% | -62.80% | -66.16% | -69.97% |
| | Cont. | $\theta \sim \mathcal{U}(1.0, 2.0)$ | -8.97% | +128.01% | -4.11% | +55.60% | -3.45% | +11.13% | -4.18% | -18.77% | -7.80% | -39.76% | -18.85% | -54.38% |
| | Cont. | $\theta \sim \mathcal{U}(1.0, 3.0)$ | -8.81% | +195.52% | -8.83% | +97.42% | -7.61% | +39.17% | -11.65% | 1.80% | -11.95% | -23.44% | -12.46% | -41.52% |
| | Disc. | $\theta \sim \mathcal{U}(\{1.0, .., 1.5\})$ | -4.67% | +75.74% | -4.21% | +16.64% | -3.46% | -17.86% | -11.82% | -40.93% | -33.11% | -57.22% | -62.68% | 68.47% |
| | Disc. | $\theta \sim \mathcal{U}(\{1.0, .., 2.0\})$ | -6.88% | +139.94% | -7.74% | +65.19% | -9.65% | +20.65% | -7.06% | -8.95% | -14.52% | -29.56% | -16.45% | -44.65% |
| | Disc. | $\theta \sim \mathcal{U}(\{1.0, .., 3.0\})$ | -13.03% | +194.21% | -12.43% | +94.99% | -10.33% | 46.73% | -13.84% | +0.37% | -14.09% | -24.56% | -13.99% | -42.05% |

■ **Dataset-Specific Adaptation Patterns.** The results reveal that optimal sampling strategies vary by dataset complexity. **KU-HAR** benefits most from moderate discrete sampling, showing clear accuracy peaks with discrete $\mathcal{U}(\{1.0, .., 2.0\})$. In contrast, **UCI-HAR** demonstrates robust performance across multiple strategies, with continuous and discrete approaches both achieving strong results. **UniMiB-SHAR** shows more conservative patterns, with moderate sampling strategies providing the best balance. Overall, we see that *ASPEN* consistently delivers substantial energy savings compared to fixed-threshold approaches. For instance, on UCI-HAR at $\theta = 1.4$, discrete sampling achieves 77.4% accuracy with 9,044 spikes, representing a 33% spike reduction and 1.6% accuracy improvement compared to the fixed-threshold baseline's best performance (75.8% accuracy and 13547 spikes). These cross-dataset results confirm that the system benefits generalize across diverse activities.

### 6.5 Neuromorphic Hardware (Xylo) Validation

To evaluate the practical impact of our approach in real-world, energy-constrained scenarios, we deployed the trained spiking models on the SynSense Xylo IMU hardware development kit (HDK). The procedures are as follows. First, we mapped our trained SNN model to the Xylo hardware using the deployment workflow provided by the Rockpool platform. This process involves quantizing the network, converting it into a compatible format, and configuring neuron parameters such as weights and thresholds for execution on the embedded SNN core. We then used a pre-encoded spike stream generated from our test dataset and streamed it directly into the SNN core, ensuring consistent and reproducible input across different energy evaluations. During inference, the digital system's core frequency was set to 50MHz, and we sampled power consumption at 5Hz, computing the average over the entire inference period.

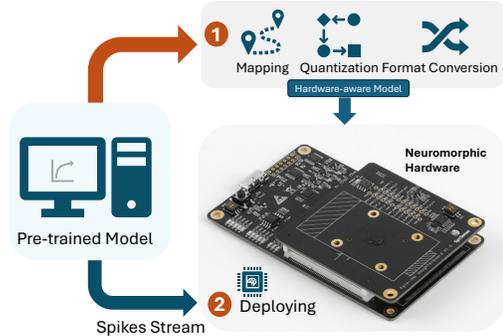

Figure 12: Deployment workflow to the Xylo IMU HDK.

Our preliminary experiments on the Xylo IMU HDK demonstrate the practical value of threshold modulation in real hardware. As expected, increasing the threshold effectively reduces energy consumption, confirming that the threshold can be used as a powerful mechanism for dynamically controlling on-chip power usage. However, while higher thresholds result in energy savings, they also lead to a significant drop in accuracy, particularly in the fixed-threshold model. In contrast, the stochastic models exhibit greater robustness, maintaining competitive accuracy even as the threshold increases. At the same accuracy level as the baseline model (fixed-threshold), energy consumption was reduced from approximately $250\mu W$ to $120\mu W$ using discrete thresholds, and to $150\mu W$ using continuous strategy. A 100 mAh, 3.7 V battery (e.g., Xiaomi Mi Band 3, Galaxy Fit, Mi Band 5) can power $120\mu W$ system for up to 128 days. A 22 mAh battery (e.g., Oura Ring) supports always-on activity recognition for 22 days. Notably, these devices do not support continuous gesture recognition and operate intermittently to save power. These results confirm that both discrete and continuous threshold modulation significantly improve energy efficiency.

## 7 CONCLUSIONS

In general, the results suggest that training spiking neural networks with moderate intrinsic noise in thresholds offers a lightweight and effective strategy for enabling energy-aware adaptation. Discrete threshold training not only preserves the energy–accuracy trade-off but also provides improved control at low spike budgets, making it particularly well-suited for ultra-low power or intermittently powered neuromorphic systems, such as batteryless or event-driven devices, where adaptability, simplicity, and reliable inference under variable energy conditions are essential.